# A Unified Initial Alignment Method of SINS Based on FGO

Hanwen Zhou, Xiufen Ye, Senior Member, IEEE

*Abstract*—The initial alignment provides an accurate attitude for SINS (strapdown inertial navigation system).

By further estimating the IMU's bias and misalignment angle, the recursive Bayesian filter is accurate. However, the prior heading error has significant influence on the convergence speed and accuracy. In addition, the accuracy will be limited by its iteration at a single time-step.

Coarse alignment method OBA (optimization-based alignment) uses MLE (maximum likelihood estimation) to find the optimal attitude quickly. However, few methods consider the IMU bias and misalignment angle, which will reduce the attitude accuracy.

In this paper, a unified method based on FGO (Factor graph optimization) and IBF (inertial base frame) is proposed. The attitude is estimated by MLE, IMU bias and misalignment angle are estimated by MAP estimation. The state of all time steps is optimized together to further improve the accuracy. Physical experiments on the rotation MEMS SINS show that the heading accuracy of this method is improved in limited alignment time.

## I. INTRODUCTION

The initial alignment process can provide the precise initial attitude for SINS, which can be achieved through process data from IMU and additional sensors like GPS[24][25]. This study focuses on initial alignment using only IMU data, which is also called self-alignment. The existing initial alignment method can be divided into three groups. 1.Coarse alignment method based on OBA and IBF. 2.Fine alignment method based on recursive Bayesian filter. 3.Two-procedure initial alignment method.

1. Coarse alignment method based on OBA and IBF

The coarse alignment provides a coarse attitude, which can be solved by SVD [2]. In recent years, the batch estimating method OBA and IBF are raised [1][4][5]. These methods are more robust to motivation by using the integral of specific force as measurement. These methods can directly determine the best initial attitude, which can be regarded as MLE, and the convergence speed is very fast. However, there are few methods considering the bias error of the IMU, which will lead to significant errors [1]. The method in [23] consider the bias estimating, but this method can only be used for stationary IMU. Method in [3] estimating the gyro's bias without accelerator's bias by adding a recursive Bayesian filter MUSQUE. However, the recursive filter will decrease the accuracy without multi-iteration.

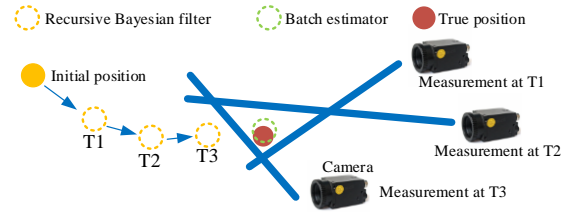

Fig. 1. MLE of Batch estimator usually converge faster than MAP estimation of recursive Bayesian filter

2. Fine alignment method based on recursive Bayesian filter

The recursive Bayesian filter is proposed by fine alignment methods. These estimators consider the priori information, which can be seen as MAP (maximum posteriori) estimation, e.g., Kalman filter or PF [6]. When the initial attitude error is large, the conventional fine alignment algorithm tends to diverge due to its strong nonlinearity and low observability [7] [8] [9]. Since then, the methods based on Lie group theory have been proposed to improve the convergence speed and the stability, such as the invariant Kalman filter [15][17] and the state transformation extended Kalman filter [16]. Along with these methods are still based on Bayesian filter, the convergence time is still long in the case of large initial attitude error. There is an example to explain why the MLE of batch estimator converges faster than MAP of recursive Bayesian filter. As shown in Fig. 1, there are three bearing measurements of feature points at three different positions. The MLE finds the optimal location directly. The estimated position of recursive Bayesian filter starts from an initial position and gradually converges to the true position. If the initial position is far from the true position, the estimated result will be worse. In addition, most recursive Bayesian filter only iterate once at a single time step, which reduces the accuracy.

3. Two-procedure initial alignment method with OBA and recursive method

The two-procedure initial alignment methods consist of the coarse alignment and the fine alignment method. First of all, coarse alignment is employed to get a coarse attitude. After a few seconds, fine alignment is used to improve the results. Since then, the BP (backward process) is raised to using these data more efficiently [26], However, the KF in this method is

This work was supported by the National Natural Science Foundation of China (Grant No. 41876100 and 42276187) and the Fundamental Research Funds for the Central Universities(Grant No. 3072022FSC0401). Hanwen Zhou, Xiufen Ye are with the College of Intelligent Systems Science and Engineering, Harbin Engineering University, Heilongjiang 150001, China., E-mail:(yexiufen@hrbeu.edu.cn). (Corresponding author: Xiufen Ye).

only optimal at single time step without multi-iteration.

To sum up, existing OBA methods has a fast convergence speed but most methods don't consider the IMU bias and misalignment angle, which will cause additional attitude error. The fine alignment method is more accurate, but its convergence speed is slow. The two-procedure initial alignment still contains a recursive Bayesian filter, only optimal at single time step without multi-iteration.

Recently, batch estimator FGO has been widely used in SLAM and integration navigation [10]-[14]. This method can estimate the associated state as a constant value or a changing sequence [18]. And method can select to estimate the state by considering or not considering the prior information. In other words, the FGO can choose to use MLE and MAP on different states. Based on this property, we can use the MLE to estimate the initial attitude at the starting moment, and use the MAP estimation to estimate the remaining states. Unlike two-procedure method solving the MLE and MAP problem in different procedure, this method solving the problems of MLE and MAP at the same time. Furthermore, the state of all time steps is optimized together to further improve the accuracy [19].

There are two main contributions in this paper:

1. Different from regular SINS state defined in the navigation coordinates frame, we choose to use the state and dynamic model in IBF, which is helpful to reduce the nonlinearity of system and make the optimization process become more stable.

2. The initial alignment problem is constructed as factor graph, the optimal value is solved by FGO, the initial consist attitude is estimated by MLE and the remaining states is estimated by MAP estimation. The state of all time steps is optimized together. This method converges fast like OBA and is more accurate than the existing method.

This paper is structured as follows. The brief introduction of different alignment methods and the structure of the proposed method are illustrated in section II. The error dynamic model and measurement model in IBF are given in section III. The unified method based on FGO is proposed in section IV. Simulations and experiments are given in sections V and VI, and the conclusions are summarized in section VII.

## II. METHOD OF COMBINING THE COARSE ALIGNMENT WITH FINE ALIGNMENT

In this section, the method of coarse alignment and fine alignment are introduced at first, and then the advantages and disadvantages of each method are analyzed. After that, the system overview of the proposed method is given out.

The nomenclature of the coordinates frames in this paper is listed in TABLE I, the different coordinates frames can also be seen in Fig. 2.

TABLE I
NOMENCLATURE

| Symbol | Description |
|---|---|
| $n$ | Ideal local level navigation coordinate frame with east-north-up geodetic axes |
| $b$ | Body coordinate frame |
| $i_{b_0}$ | Nonrotating initial body coordinate frame at start time of initial alignment |
| $\tilde{i}_{b_0}$ | The calculated initial body coordinate frame at the start time of initial alignment |
| $i_{n_0}$ | Nonrotating initial navigation coordinate frame at start time of initial alignment |
| $i$ | Nonrotating earth center coordinate frame |
| $e$ | Earth coordinate frame |

As shown in Fig. 2, $i$ is the earth center coordinate frame. $i_{b_0}$ is an inertial frame, which is coincides with the body coordinate frame at the start time. $i_{n_0}$ is an inertial frame coinciding with a local navigation frame at the start time. $g_n$ indicates the gravity in $n$ frame.

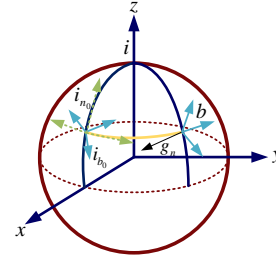

Fig. 2. Coordinates frame definition

### A. Coarse Alignment Based on OBA and IBF

The OBA method divides initial alignment into two parts, one part is attitude tracking, which is responsible for updating DCM (direction cousin matrix) $C_b^{i_{b_0}}$, and the other part is responsible for determining the constant initial attitude $C_{i_{b_0}}^{i_{n_0}}$. Since $C_{i_{n_0}}^n = C_n^{i_{n_0}'}$ is easy to be obtained through (4), the output DCM $C_b^n$ can be collected through $C_b^n = C_{i_{n_0}}^n C_{i_{b_0}}^{i_{n_0}} C_b^{i_{b_0}}$.

1. Attitude track in inertial frame—calculating the $C_b^{i_{b_0}}$

Differential equation of DCM can be written as (1):

$$\dot{C}_b^{i_{b_0}} = C_b^{i_{b_0}} (\omega_{i_{b_0}b}^{b \wedge}) \tag{1}$$

where, $C_b^{i_{b_0}}$ can be expressed as (2):

$$\begin{aligned} C_b^{i_{b_0}} &= C_{b_{k-1}}^{i_{b_0}} C_b^{b_{k-1}} \\ C_b^{b_{k-1}} &= \exp\left(\left(T\omega_{ib}^i\right)^\wedge\right) \end{aligned} \tag{2}$$

The exp(.) represents the matrix exponential, (.)^ represents linear, skew-symmetric operator. DCM $C_{b_{k-1}}^{i_{b_0}}$ represents the $C_b^{i_{b_0}}$ before time interval $T$. This DCM can be calculated through Rodriguez formulation [19].

The attitude track doesn't consider the bias error of IMU, the bias error of $\omega_{ib}^i$ will cause the attitude error to increase continuously.

2. Constant initial attitude determination—calculating the $C_{i_{n_0}}^{i_{b_0}}$.

The integration of specific forces in inertial frame $i_{b_0}$ and $i_{n_0}$ can be written as (3):

$$F_i^{i_{b_0}} = \int_0^{kdt} C_b^{i_{b_0}} f_{sf}^b$$
$$G_i^{i_{n_0}} = -\int_0^{kdt} g_{i_{n_0}} \qquad (3)$$

Where the $C_{i_{n_0}}^{i_{b_0}}$ represents the constant initial attitude, $f_{sf}^b$ is the specific force measurement. $g_{i_{n_0}}$, which is the projection of gravity in $i_{n_0}$ can be expressed as:

$$g_{i_{n_0}} = C_n^{i_{n_0}} g_n$$
$$C_n^{i_{n_0}} = \exp\left((T\omega_{ie}^n)^\wedge\right) = \begin{bmatrix} \cos\omega_{ie}t & -\sin\omega_{ie}t\sin L & \sin\omega_{ie}t\cos L \\ \sin\omega_{ie}t\sin L & 1-(1-\cos\omega_{ie}t)\sin^2 L & (1-\cos\omega_{ie}t)\sin L\cos L \\ -\sin\omega_{ie}t\cos L & (1-\cos\omega_{ie}t)\sin L\cos L & 1-(1-\cos\omega_{ie}t)\cos^2 L \end{bmatrix} \qquad (4)$$

Where $L$ represents the latitude, $\omega_{ie}^i = [0\ 0\ \omega_{ie}]^T$ indicates the rotation rate of the earth and $g_n = [0\ 0\ -g]^T$ indicates the gravity vector. Then the relationship between $F_i^{i_{b_0}}$ and $G_i^{i_{n_0}}$ can be expressed as (5).

$$C_{i_{b_0}}^{i_{n_0}} F_i^{i_{b_0}} = G_i^{i_{n_0}} + \int_0^{kdt} C_{i_{b_0}}^{i_{n_0}} \hat{\nabla}^{b_0}$$
$$C_{i_{b_0}}^{i_{n_0}} F_i^{i_{b_0}} \approx G_i^{i_{n_0}} \qquad (5)$$

Where $\hat{\nabla}^{b_0}$ represents the disturbance acceleration, $G_i^{i_{n_0}}$ is the integration of gravity. The cost function for OBA can be expressed as (6).

$$C_{i_{b_0}}^{i_{n_0}*} = \min\left(\sum_{k=1}^n \left\| C_{i_{b_0}}^{i_{n_0}} F_i^{i_{b_0}} - G_i^{i_{n_0}} \right\|\right) \qquad (6)$$

The optimal $C_{i_{b_0}}^{i_{n_0}}$ can be obtained directly by minimum this function, then $C_b^n = C_{i_{n_0}}^n C_{i_{b_0}}^{i_{n_0}} C_b^{i_{b_0}}$ can be obtained, and it will not be influenced by initial attitude and can converge rapidly.

*B. Fine Alignment Based on Recursive Bayesian Filter*

The fine alignment models the SINS error as system state $x$:

$$x = \begin{bmatrix} \phi & \delta v & \varepsilon & \nabla \end{bmatrix} \qquad (7)$$

Where $\phi$ and $\delta v$ represent the misalignment angles and the velocity errors defined in the navigation frame, $\varepsilon$ and $\nabla$ represent the gyro bias errors and accelerometer bias errors respectively. With the error SINS dynamic model and measurement equation, the Bayesian filter, e.g., KF can be employed to estimate the system state. As a MAP estimator, the Bayesian filter must set a prior information, and optimal at single time step owing to the Markov assumption and the recursive property.

*C. The Unified Initial Alignment Method Base on FGO*

The main difference between MAP and MLE is that the MAP considers the prior information. The prior information expressed by the prior factor in factor graph, we can choose not to set prior factor for associated state to estimate them by MLE.

The proposed FGO estimates constant initial attitude $C_{i_{n_0}}^{i_{b_0}}$ by MLE, and estimates the misalignment angle and the bias of IMU by MAP. Besides, the FGO optimal the whole trajectory with multi-iteration.

*D. System Overview of the Unified Initial Alignment Method*

The structure of the proposed method is shown in Fig. 3, and the corresponding procedures are given as follows.

1. Calculating the result of attitude track, integration of specific force and gravity, the equation is illustrated in A of section II.

2. Combining the system state model and the measurement model (section III), and construct a factor graph to represent the relationship between measurements and states at different times. By minimizing the cost function, we can solve the optimal value. This is given in section IV.

3. After obtaining the optimal misalignment angle and constant initial attitude, the optimal attitude output can be calculated by equation (24).

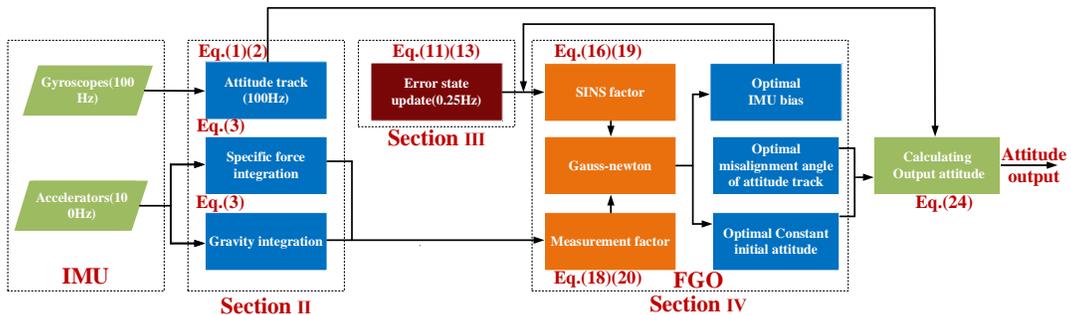

Fig. 3 Structure of the proposed method

## III. SINS Model and Measurement Model in IBF

There are two types of nodes in factor graph, variables and factors. When using factor graph, the system state, the system dynamic model and the measurement model should be determined at first. The system states are constructed as variables, measurement model and system model are constructed as factors.

The regular system model of SINS defined in navigation coordinates frame will lead to diverge of batch estimator, the dynamic model of SINS and the measurement model defined in IBF is illustrated in this section.

### A. The Nonlinear Property of Regular System Dynamic Model

The Rodrigues's Formula can be expressed as:

$$\exp\left((\phi)^\wedge\right) = I + \frac{\sin\phi}{\phi}(\phi)^\wedge + \frac{1-\cos\phi}{\phi^2}\left((\phi)^\wedge\right)^2 \quad (8)$$

This is nonlinear function, during the derivation of error dynamic model of SINS, a simplify property of Rodrigues's Formula $\exp\left((\phi)^\wedge\right) \approx I + (\phi)^\wedge$ is employed. However, this property can only establish when $\phi$ is small. The $\phi$ defined in navigation coordinate frame could be large (0 to 180 degree), which makes the system becoming highly nonlinear, resulting in the divergence of batch estimator or recursive estimator.

### B. Using the System State in IBF With Better Linear Property

We choose the state defined in IBF for the proposed method, the corresponding state is:

$$x_k = \begin{bmatrix} \phi_k & \delta F_i^{i_{b_0}}(k) & \varepsilon_k & \nabla_k \end{bmatrix} \quad (9)$$

It is necessary to clarify exactly that the state is different with state in regular fine alignment (7), and the $\phi_k$ and $\delta F_i^{i_{b_0}}(k)$ are defined in inertial coordinates frame at start time, $i_{b_0}$, but not in navigation frame. $\phi_k$ indicates the misalignment angle in $i_{b_0}$, the $\phi_k$ is always small because it grows from zero. The small $\phi_k$ guaranteed the establish of the simplify property on Rodrigues's Formula $\exp\left((\phi)^\wedge\right) \approx I + (\phi)^\wedge$ and make the estimating procedure become more stability.

$\delta F_i^{i_{b_0}}(k)$ represents the error of specific force integration. $\varepsilon_k$ represents the gyro's bias error, $\nabla_k$ is the accelerator's bias error. Remarkably, the chosen states of attitude track $\phi_k$ and $\delta F_i^{i_{b_0}}(k)$ are the error states, this is very important for factor graph expressed by fewer factors. The error state will not change fast, therefore, the time interval between the adjacent factors can be set longer, which reduces the number of factors.

The SINS states at different times and the constant initial attitude $C_{i_{b_0}}^{i_{b_0}}$ are added into system state, the state can be expressed as (10). The state will be transformed to the variables of factor graph.

$$X = \begin{pmatrix} x_1 & x_2 & x_3 & \cdots & x_n & C_{i_{b_0}}^{i_{b_0}} \end{pmatrix} \quad (10)$$

### C. The Dynamic Model of SINS in IBF

The attitude error differential equation of SINS [26] is expressed as (11).

$$\dot{\phi} = -C_b^{i_{b_0}} \delta\omega_{i_{b_0}b}^b \quad (11)$$

Then we derive the dynamic model for the error specific force integration $\delta F_i^{b_0}$.

$$\delta F_i^{b_0} = \tilde{F}_i^{b_0} - F_i^{b_0} \quad (12)$$

Define $\tilde{f}_{sf}^b = f_{sf}^b + \delta f_{sf}^b$ as the specific force measurement in $b$, where $\delta f_{sf}^b$ represents the measurement error. Differential on both sides of the (12), the equation can be further derived as (13).

$$\begin{aligned}
\delta \dot{F}_i^{b_0} &= C_b^{\tilde{i}_{b_0}} \tilde{f}_{sf}^b - C_b^{b_0} f_{sf}^b \\
&\approx C_b^{\tilde{i}_{b_0}}\left(f_{sf}^b + \delta f_{sf}^b\right) - \left[I + (\phi^\wedge)\right] C_b^{\tilde{i}_{b_0}} f_{sf}^b \\
&= C_b^{\tilde{i}_{b_0}} f_{sf}^b + C_b^{\tilde{i}_{b_0}} \delta f_{sf}^b - C_b^{\tilde{i}_{b_0}} f_{sf}^b - (\phi^\wedge) C_b^{\tilde{i}_{b_0}} f_{sf}^b \\
&= C_b^{\tilde{i}_{b_0}} \delta f_{sf}^b - (\phi^\wedge) C_b^{\tilde{i}_{b_0}} f_{sf}^b \\
&= C_b^{\tilde{i}_{b_0}} \delta f_{sf}^b + \left(C_b^{\tilde{i}_{b_0}} f_{sf}^b\right)^\wedge \phi
\end{aligned} \quad (13)$$

Since the bias of the IMU does not change in a short time, the dynamic model of SINS can be rewritten as (14).

$$\begin{aligned}
\dot{\phi} &= -C_b^{b_0} \varepsilon \\
\delta \dot{F}_i^{i_{b_0}} &= C_b^{\tilde{i}_{b_0}} \nabla + \left(C_b^{\tilde{i}_{b_0}} f_{sf}^b\right)^\wedge \phi \\
\dot{\varepsilon} &= 0 \\
\dot{\nabla} &= 0
\end{aligned} \quad (14)$$

And the INS model can be defined as (15).

$$x_{k+1} = f(x_k) + N\left(0, \Sigma_k^{INS}\right) \quad (15)$$

Where $f(x_k)$ can be expressed as (16), which is obtained by the Eulerian method in the SINS dynamic model (14). And $dt$ is the time interval, which was set as two seconds in the simulations and experiments.

$$f(x_k) = \begin{bmatrix} \phi_k - C_b^{\tilde{i}_{b_0}} \varepsilon_k dt \\ \delta F_i^{i_{b_0}}(k) + \left(C_b^{\tilde{i}_{b_0}} \nabla_k + \left(C_b^{\tilde{i}_{b_0}} f_{sf}^b\right)^\wedge \phi_k\right) dt \\ \varepsilon_k \\ \nabla_k \end{bmatrix} \quad (16)$$

$N\left(0, \Sigma_k^{INS}\right)$ indicates the Gauss noise and represents the noise of the system state.

### D. Measurement Model of SINS in Attitude Track

The measurement model can be expressed as (17).

$$z_k = h(x_k, C_{i_{n_0}}^{i_{b_0}}) + N(0, \Sigma_k^{IOSF}) \quad (17)$$

where $h(x_k, C_{i_{n_0}}^{i_{b_0}})$ is expressed as:

$$h(x_k, C_{i_{n_0}}^{i_{b_0}}) = -C_{i_{n_0}}^{i_{b_0}} \int_0^k g^{n_0} dt - \left(F_i^{i_{b_0}}(k) - \delta F_i^{i_{b_0}}(k)\right) \quad (18)$$

And $N(0, \Sigma_k^{IOSF})$ representing the noise of measurement.

## IV. Unified Initial Alignment Method of SINS Based on FGO

In section III, the dynamic model and the measurement model in IBF have been illustrated. In this section, we construct a unified initial alignment method based on FGO.

### A. Formulations

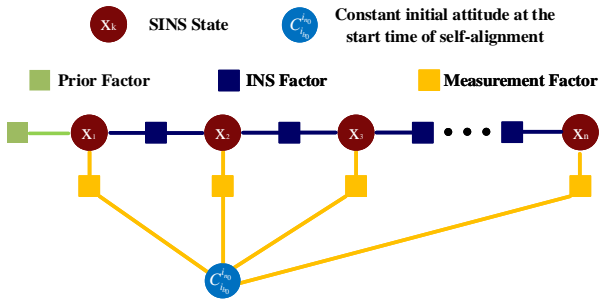

Fig. 4 Factor graph structure of the improved method

Fig. 4 shows the factor graph structure of the proposed method, in which the SINS state is represented by circles, and the factors are represented by squares, set as follows:

1. INS factor

The error function of INS factor can be expressed as (19).

$$\left\| e_k^{INS} \right\|_{\Sigma_k^{INS}}^2 = \left\| x_{k+1} - f(x_k) \right\|_{\Sigma_k^{INS}}^2 \quad (19)$$

2. Measurement factor

The error function of measurement factor can be expressed as (20).

$$\left\| e_k^{IOSF} \right\|_{\Sigma_k^{IOSF}}^2 = \left\| z_k - h(x_k, C_{i_{n_0}}^{i_{b_0}}) \right\|_{\Sigma_k^{IOSF}}^2 \quad (20)$$

3. Prior factor

The error function of prior factor can be expressed as (21).

$$\left\| e^{PRIOR} \right\|_{\Sigma^{PRIOR}}^2 = \left\| r_{PRIOR} - H_{PRIOR} x_1 \right\|_{\Sigma^{PRIOR}}^2$$
$$r_{PRIOR} = 0_{12 \times 12}$$
$$H_{PRIOR} = I_{12 \times 12} \quad (21)$$

With a suitable prior information, the estimation of states will be smoother in beginning. It is necessary to clarify that this prior factor not set for the $C_{i_{b_0}}^{i_{n_0}}$. The estimation of $C_{i_{b_0}}^{i_{n_0}}$ can still be seen as MLE. Where $\Sigma^{PRIOR}$ is diagonal matrix, which represents the noise of initial state. The $r_{PRIOR} = 0_{12 \times 12}$ is very suitable, because the misalignment angle and the specific error are zero at beginning, and the gyro's bias is very small. This prior factor will not influence the results, and the prior factor can also be deleted if the smoothness at beginning is not important.

### B. Optimization and Jacobian Calculation

The FGO can be transformed into minimizing a quadratic function[20][21].

$$X^* = \arg\min \sum_k \left\| e_k^{INS} \right\|_{\Sigma_k^{INS}}^2 + \left\| e_k^{IOSF} \right\|_{\Sigma_k^{IOSF}}^2 + \left\| e^{PRIOP} \right\|_{\Sigma^{PRIOR}}^2 \quad (22)$$

$$\frac{\partial h(x_k, C_{i_{n_0}}^{i_{b_0}})}{\partial C_{i_{n_0}}^{i_{b_0}}} = \lim_{\varphi \to 0} \frac{-\exp(\varphi^\wedge)\exp(\xi^\wedge)\int_0^k g^{n_0} dt + \exp(\xi^\wedge)\int_0^k g^{n_0} dt}{\varphi}$$
$$\approx \lim_{\varphi \to 0} \frac{-(I + \varphi^\wedge)\exp(\xi^\wedge)\int_0^k g^{n_0} dt + \exp(\xi^\wedge)\int_0^k g^{n_0} dt}{\varphi}$$
$$= \lim_{\varphi \to 0} \frac{-(\varphi^\wedge)\exp(\xi^\wedge)\int_0^k g^{n_0} dt}{\varphi}$$
$$= \lim_{\varphi \to 0} \frac{\left(\exp(\xi^\wedge)\int_0^k g^{n_0} dt\right)^\wedge \varphi}{\varphi}$$
$$= \left(C_{i_{n_0}}^{i_{b_0}} \int_0^k g^{n_0} dt\right)^\wedge \quad (23)$$

Then, the Gauss-Newton method is used to minimize this function. To calculate the Jacobian matrix, the perturbation $\varphi$ is applied on the left respect to DCM [19], then the Jacobian matrix can be calculated as (23).

### C. Optimal attitude output

After getting the optimal $C_{i_{b_0}}^{i_{n_0}}$ and $C_{\tilde{i}_{b_0}}^{i_{b_0}}$, the optimal output DCM can be expressed as (24).

$$C_b^n = C_{i_{n_0}}^n C_{i_{b_0}}^{i_{n_0}} C_{\tilde{i}_{b_0}}^{i_{b_0}} C_b^{\tilde{i}_{b_0}} \quad (24)$$

## V. Simulation

### A. Simulation Setup

The simulation compared the heading results of five different methods, the first one is the conventional two procedure method OBA with Kalman filter[6], in which the OBA method works for the first 60s, followed by KF; the second one is the recursive method ST-EKF [16], the third one is the OBA method [4]. The fourth method is the OBA method with KF and BP (backward process) [26], in order to get the heading error changing with time, we run this method every 20 seconds, save the results of last 20 seconds, and put the results together. The last one is the method proposed in this paper.

Simulation tests were set up under mooring conditions, bias error are added into the raw IMU data, where gyro bias are set as $\varepsilon_x^b = -8°/h$, $\varepsilon_y^b = 6°/h$, $\varepsilon_z^b = -7°/h$, bias of accelerator are set as $\nabla_x^b = 1mg$, $\nabla_y^b = -1mg$, $\nabla_z^b = 1mg$. The IMU in simulation and experiment will continuously rotated [22] to improve the observability of the misalignment angle and bias of gyros, and all these states will become observable. Each method was tested for 10 minutes, and RMSE (Root Mean Square Error) was chosen as the evaluation criterion.

## B. Simulation Results and Analysis

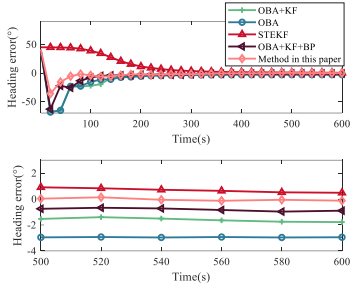

Fig. 5 Heading error of different methods

As shown in Fig. 5, The upper part of the diagram shows the heading error of 600 seconds, the proposed method and the OBA method based on batch estimator converge to zero very fast. The method with BP also converges very fast by three times forward KF and backward process, the influence of prior information is reduced.

TABLE II
HEADING RMSE OF DIFFERENT INITIAL ALIGNMENT METHODS

| Average heading RMSE | OBA+ KF | OBA | STEKF | OBA+KF +BP | Method in this paper |
|---|---|---|---|---|---|
| 50-100 sec | 25.268 | 23.128 | 42.784 | 20.015 | **8.028** |
| 150-200 sec | 5.436 | 7.000 | 17.564 | **3.635** | 3.882 |
| 250-300 sec | 3.577 | 3.821 | 4.583 | 2.857 | **2.232** |
| 550-600 sec | 1.684 | 2.947 | 0.571 | 0.872 | **0.104** |

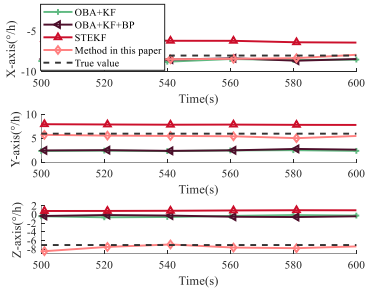

Fig. 6 Estimated results of Gyro bias

The heading RMSE in different times is shown in TABLE II. Between 50 and 200 seconds, the RMSE of the OBA method and the proposed method is better than that of STEKF and KF, because the batch estimation converges faster than the Bayesian filter. Between 250 and 600 seconds, the heading RMSE of KF and the proposed method is smaller than OBA, because they estimate the gyro's bias and misalignment angle. As shown in Fig. 6, the proposed method has the best bias estimation effect for gyros, especially Z-axis gyro. The heading RMSE of the proposed method is 0.104 between 550 and 600 seconds, and its performance is still better than other methods after 250 seconds. There are two reasons why the proposed method has the best accuracy:

1. The MLE of the batch estimator converges more rapidly than the MAP of recursive Bayesian filter in initial alignment.
2. In each iteration of FGO, a re-linearization and optimal on the state of all time steps is performed. The multiple re-linearization can effectively mitigate the linearization error, and the optimal on whole trajectory will improve the accuracy.

To sum up, there are two conclusions.

1. By estimate the $C_{i_{b_0}}^{i_{n_0}}$ like MLE and estimate the other states as MAP, the proposed method based on FGO has a faster converge speed than recursive method STEKF or KF.

2. With the same alignment time, the proposed method is more accurate than existing OBA method through estimating the IMU bias and misalignment angle. And the method also accurate than other method with the rapid converge speed and the multiple optimize on the state of all time steps.

## VI. PHYSICAL EXPERIMENT

### A. Physical Experiment Setup

Dual-antenna receivers were set as attitude references. As shown in Fig. 7, GPS antennas were set up in the fore and aft of a ship, and the length of the baseline was 15 meters. Thus, the accuracy of the attitude reference was about $0.2°/15 \approx 0.013°$. The system was fixed on the deck of the ship. All of the connections were tight.

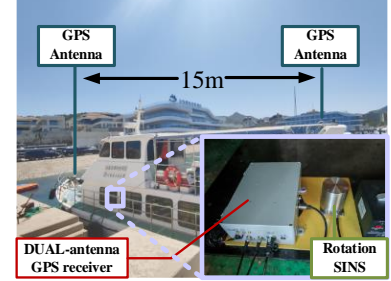

Fig. 7. Experiment ship and experiment system

TABLE III
PERFORMANCE OF THE ROTATIONAL SINS

| Parameters | Value |
|---|---|
| Gyro bias repeatability | 2.00°/h |
| Gyro bias stability | 0.1°/h |
| Gyro random walk | 0.1°/√h |
| Gyro scale factor accuracy | 100ppm |
| Accelerator bias repeatability | 20.0ug |

The performance parameters of the experiment rotation SINS are shown in TABLE III. The initial alignment experiments were carried out 5 times, 5 minutes each time, and RMSE was selected as the evaluation criteria.

### B. Experimental Results

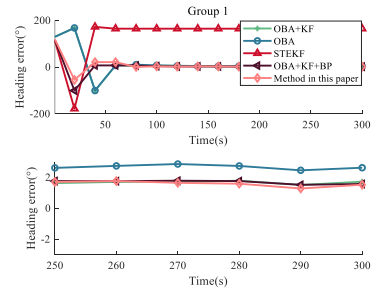

Fig. 8 Heading error comparison of group 1

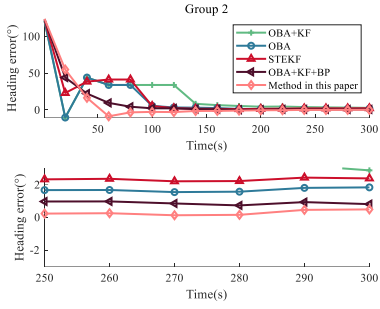

Fig. 9 Heading error comparison of group 2

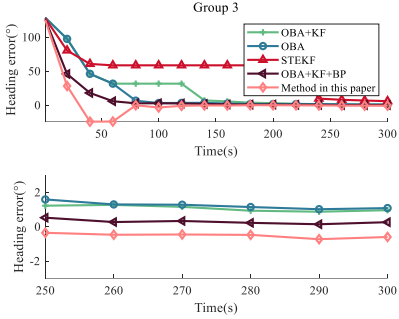

Fig. 10 Heading error comparison of group 3

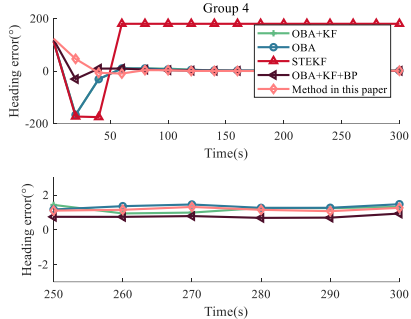

Fig. 11 Heading error comparison of group 4

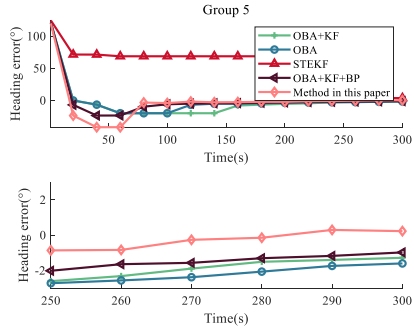

Fig. 12 Heading error comparison of group 5

The upper half of Fig. 8 to Fig. 12 is the heading error among 300 seconds, and the bottom half of the figures is the heading error among 250 to 300 seconds.

From these figures, it can be seen that the OBA method, the method with BP and the proposed method converged rapidly, and the detailed comparison on the average RMSE of the five groups is given in TABLE IV.

TABLE IV
HEADING RMSE OF DIFFERENT INITIAL ALIGNMENT METHODS

| Average heading RMSE | OBA+KF | OBA | STEKF (group2,3,5) | OBA+KF+BP | Method in this paper |
|---|---|---|---|---|---|
| 50-100 sec | 20.622 | 17.512 | 54.145 | **10.983** | 16.888 |
| 150-200 sec | 4.264 | 2.820 | 38.538, | 1.869 | **1.317** |
| 250-300 sec | 1.872 | 1.840 | 4.557 | 1.060 | **0.819** |

Between 50 and 200 seconds, the accuracy of the OBA method and the proposed method is better than that of STEKF and KF, the batch estimation converges faster than the Bayesian filter. The method with BP also converges very fast by three times forward KF and backward process to reduce the influence of prior information.

Between 250 and 300 seconds, by estimating the misalignment angle and the bias, the heading RMSE of the proposed method and method with BP is less than that of OBA. STEKF has the largest heading error, which also proves that the initial attitude error has a significant influence on the recursive filter.

Although the RMSE differences between different methods decreases with the alignment time, between 150 and 300 seconds, the improved method still better than other method in heading RMSE.

To sum up, there are some conclusions.

1. The initial attitude significantly influences the performance of the recursive filter. These filters may not converge to the true value if the prior information have a large error. The proposed method and the OBA method estimate the initial consist attitude by MLE, which will not be influenced by prior information and converge faster if the state has low observability.

2. This method estimates the bias of IMU and misalignment angle as MAP and which is more accurate than OBA method, because the OBA don't consider the IMU bias and misalignment angle.

3. With the same alignment time, the proposed method is more accurate than other method with the rapid converge speed and the multiple optimize on the state of all time steps.

## VII. CONCLUSION

This paper proposed a novel unified initial alignment method of SINS based on FGO. Firstly, we proposed the error dynamic model and measurement model of IBF to reduce the nonlinear property of the system. Then, the corresponding factor graph is constructed, and finally the initial constant attitude, the misalignment angle and IMU bias are estimated through FGO.

The FGO can achieve MLE and MAP estimation for different states on the same time. By batch estimating the optimal attitude by MLE, estimating the IMU bias and the misalignment angle by MAP estimation simultaneously, this unified method successfully combines fast convergence of MLE and high accuracy of MAP estimator.

With the same initial alignment time, the proposed method is more accurate than KF and STEKF or the two-procedure initial alignment method with the rapid converge speed and the multiple optimize on the state of all time steps. Therefore, by using this improved method, higher accuracy can be obtained in a shorter time.

The method can be extended to other state estimation systems. If the prior information may have large error and the associate state has low observability, by using FGO, we can estimate this state by MLE to eliminate the influence of initial value. Such as lever arm and time drift of GPS/SINS navigation system.